\documentclass[letterpaper, 10 pt, conference]{ieeeconf}  

\IEEEoverridecommandlockouts                              

\usepackage{graphics} 
\usepackage{epsfig} 
\usepackage{mathptmx} 
\usepackage{times} 
\usepackage{amsmath} 
\usepackage{amssymb}  
\usepackage{tabu}                      
\usepackage{booktabs}                  
\usepackage{multirow}
\usepackage{bm}
\usepackage{cite}
\usepackage{hyperref}

\title{\LARGE \bf
iDF-SLAM: End-to-End RGB-D SLAM with Neural Implicit Mapping and Deep Feature Tracking
}

\author{Yuhang Ming$^{1}$, Weicai Ye$^{2}$ and Andrew Calway$^{1}$
\thanks{$^{1}$Yuhang Ming and Andrew Calway are with the Visual Information Laboratory, Department of Computer Science, University of Bristol, Bristol, U.K.
        {\tt\small \{yuhang.ming, andrew.calway\}@bristol.ac.uk}}%
\thanks{$^{2}$Weicai Ye is with the State Key Lab of CAD\&CG, Zhejiang University, Hangzhou, China.
        {\tt\small weicaiye@zju.edu.cn}}%
}

\begin{document}

\maketitle
\thispagestyle{empty}
\pagestyle{empty}

\begin{abstract}

We propose a novel end-to-end RGB-D SLAM, iDF-SLAM, which adopts a feature-based deep neural tracker as the front-end and a NeRF-style neural implicit mapper as the back-end. The neural implicit mapper is trained on-the-fly, while though the neural tracker is pretrained on the ScanNet dataset, it is also finetuned along with the training of the neural implicit mapper. Under such a design, our iDF-SLAM is capable of learning to use scene-specific features for camera tracking, thus enabling lifelong learning of the SLAM system. Both the training for the tracker and the mapper are self-supervised without introducing ground truth poses. We test the performance of our iDF-SLAM on the Replica and ScanNet datasets and compare the results to the two recent NeRF-based neural SLAM systems. The proposed iDF-SLAM demonstrates state-of-the-art results in terms of scene reconstruction and competitive performance in camera tracking. 

\end{abstract}

\section{Introduction}

RGB-D simultaneous localisation and mapping (SLAM) aims to estimate the trajectory of a moving RGB-D camera while performing reconstruction of the surrounding environment. It is a vital ability to the applications such as robots, autonomous vehicles and virtual/augmented realities to ensure robust operations.

Depending on the constructed maps, conventional RGB-D SLAM systems are commonly categorised into sparse SLAM ~\cite{orbslam, dai2022}
, dense SLAM \cite{kinectfusion, elasticfusion, dvo} 
and hybrid ones \cite{bundlefusion, fdslam}. 
Later, with advances in deep neural networks (DNNs), much attention has been paid to improve various aspects of SLAM with semantic information extracted with DNNs, such as meaningful mapping \cite{meaningfulmap, semanticfusion, PVO}, dynamic tracking \cite{quadricrgbdslam, objslam, DeFlowSLAM}, relocalisation \cite{objreloc, semanticpr}, \textit{etc.} Although these systems have demonstrated promising performance in terms of both tracking and reconstruction accuracy, they suffer from the huge video memory (VRAM) footprint in storing the reconstructed map and the computational consumption when modifying the map on-the-fly.

With the focus on exploring a more efficient map representation, CodeSLAM \cite{codeslam} took the idea of learnt depth from images and presented a compact yet dense representation of the map with optimisable vectors learnt in an auto-encoder architecture. Its success has inspired many successors \cite{scenecode, deepfactors, codemapping} but the requirement of pretraining makes them difficult to generalise to new environments.
More recently, the success of NeRF \cite{nerf} proves the fact that multi-layer perceptrons (MLPs) are capable of encoding scene geometries in fine detail. Adopting this as the map representation in the SLAM, iMAP \cite{imap} and NICE-SLAM \cite{niceslam} are published successively to train the neural implicit network on-the-fly and achieve astonishing results.
However, when estimating poses, both methods performed inverse NeRF optimisation on every frame, which is very time-consuming.

\begin{figure*}[h!]
    \centering
    \includegraphics[width=\textwidth]{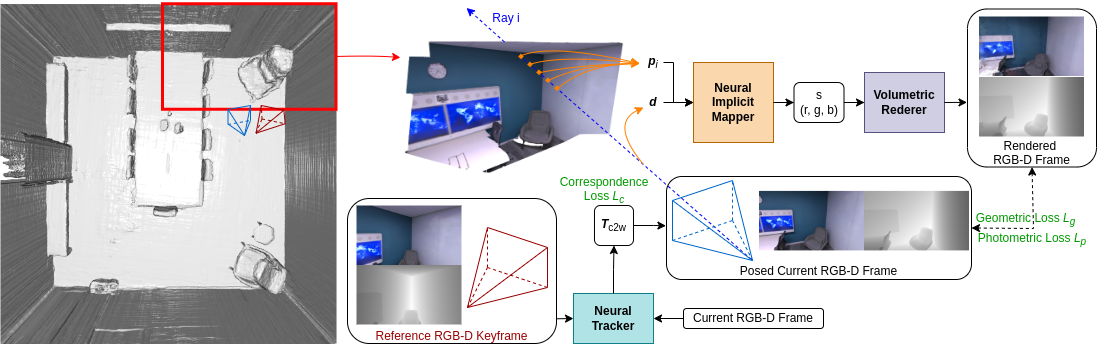}
    \caption{An overview of the data flow in the proposed iDF-SLAM.}
    \label{fig:overview-dataflow}
    \vspace{-3ex}
\end{figure*}

In this paper, we focus on improving the runtime performance of NeRF-based SLAMs.
Inspired by FD-SLAM \cite{fdslam}, which combines feature-based tracking with dense mapping, we propose iDF-SLAM which uses a feature-based neural tracker as the front-end to track the camera trajectory and adopts a NeRF-style neural implicit network as the map representation in the back-end. An alternative pose and map updating scheme is adopted in our iDF-SLAM and the map is only updated when a new keyframe is needed.
An overview of the proposed system is shown in Fig. \ref{fig:overview-dataflow}. 

For the neural tracker, we take advantage of recent advances in the pairwise point cloud registration network and modified it to make it work robustly as a SLAM front-end. Although the neural tracker is pretrained on the public dataset, we propose to finetune the tracker network as the system is running to enable it to learn the best scene-specific features for camera pose estimation, thus empowering the lifelong learning ability of the SLAM system.

Regarding the neural implicit mapper, as in iMAP \cite{imap}, we adopt a single MLP to represent the map. To obtain a better reconstructed map, we propose to let the MLP directly regress the truncated signed distance function (T-SDF) at a given 3-D position, rather than volume density in iMAP \cite{imap} or occupancy in NICE-SLAM~\cite{niceslam}. The MLP is initialised with the first frame of the sequence. Whenever an input frame is selected as the new keyframe, we trigger the map updating process by first optimising the pose with respect to the latest map and then updating the weights of the MLP given the keyframe with the optimised pose. Extensive experiments on Replica and ScanNet datasets demonstrate that our iDF-SLAM achieves SOTA scene reconstruction results and competitive performance in camera tracking.

\section{Related Work}

\textbf{Reconstruction with neural implicit networks}: Given a set of posed images, NeRF \cite{nerf} was proposed with the focus on novel view synthesis. It represents the scene with an MLP and renders the scene with predicted volume densities. Although the volume density representation is good for transparent objects, it suffers from inaccurate depth estimation, thus leading to suboptimal surface prediction for reconstruction tasks. Improving on this, UNISURF \cite{unisurf} proposed to unify volume and surface rendering by iterative root-finding. By replacing the volume density with SDF, NeuS \cite{neus} was proposed to directly regress the SDF value along with a new rendering formulation. Incorporating additional depth measurements, \cite{neuralrgbd} presented a hybrid scene representation with a single MLP to achieve high-quality reconstruction. Working in a different direction, NSVF \cite{nsvf} introduced the sparse voxel tree to model local geometry with implicit fields bounded by voxels. Similarly, NeuralBlox \cite{neuralblox} represents the scene as an occupancy grid with latent codes stored in each cell and updated incrementally. In this work, we do not use the voxel representation and let the network directly regress the T-SDF values.

\textbf{Poses estimation with NeRF}: Focusing on pose estimation, iNeRF \cite{inerf} and NeRF$--$ \cite{nerf--} are concurrent work and pioneers in estimating poses of a given image when the neural implicit network is fully trained. This goal is achieved by ``inverting'' the NeRF optimisation and it is proved that the performance of the neural implicit network can be further improved with inverse NeRF optimisation. Pushing it to the limits, BARF \cite{barf} proposed to train the neural implicit network using images with inaccurate and even unknown poses through bundle adjusting. Though this idea is very close to NeRF-based SLAM, the main difference is that BARF takes in a set of images simultaneously, while in SLAM problems, the images are input sequentially.

\textbf{SLAM based on NeRF:} Finally, there are few works that use neural implicit networks as the only map representation in a SLAM system, iMAP \cite{imap} and NICE-SLAM \cite{niceslam}. These are the closest works to ours, but the proposed iDF-SLAM differs from them in the following ways. With the introduction of a neural tracker, our iDF-SLAM is capable of updating the map at a much lower frequency, thus improving runtime performance. Compared to iMAP, our iDF-SLAM uses SDF representation, leading to better reconstruction results. Compared to NICE-SLAM, which uses four different MLPs in mapping, we use only a single MLP to represent the map.

\section{System Overview}
\begin{figure}[t]
    \centering
    \includegraphics[width=0.48\textwidth]{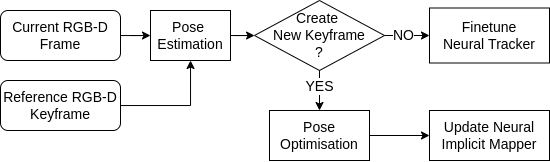}
    \caption{System pipeline overview of the proposed iDF-SLAM.}
    \label{fig:overview-system}
    \vspace{-3ex}
\end{figure}

The proposed iDF-SLAM consists of three main components, namely \textit{neural tracker}, \textit{neural implicit mapper} and \textit{volume renderer}. 
An overview of the proposed iDF-SLAM is shown in Fig. \ref{fig:overview-system} with its input being a sequence of RGB-D frames. 
At each timestep, a current RGB-D frame $F_c = \{ I_c, D_c \}$ and a reference RGB-D keyframe $KF_r = \{ I_r, D_r \}$ are fed into the neural tracker to estimate the pose of the current frame $\bm{T}_c$, where $I$ and $D$ are the corresponding RGB image and depth map. 
Then, the current frame is chosen as the new keyframe if there is enough motion with respect to the last keyframe or
the tracker gets lost on the current frame.
If a new keyframe is created at current timestep, pose optimisation is first performed with the weights of the neural implicit mapper fixed. Then, using this optimised pose of the current frame, the weights in the neural implicit mapper are optimised. 
However, if the current frame is not selected as the new keyframe, we fix the weights of the neural implicit mapper and perform finetuning on the neural tracker. 
Finally, the volumetric renderer is used to generate rendered RGB-D rays/frames for the loss computation and visualisation purposes. 

Since we use exactly the same rendering approach as \cite{neuralrgbd} in our volumetric renderer, we refer the readers to their paper for details. 
In the following sections, we will discuss the neural implicit mapper and the neural tracker of our proposed iDF-SLAM in detail.

\section{Neural Implicit Mapper}

\begin{figure}[t]
    \centering
    \includegraphics[width=0.5\textwidth]{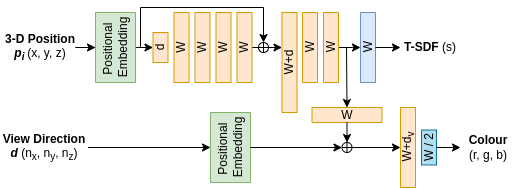}
    \caption{Architecture of the MLP in the neural implicit mapper.}
    \label{fig:mapper}
    \vspace{-3ex}
\end{figure}

Similar to iMAP \cite{imap}, the core of our neural implicit mapper is a single MLP, which is used as the only map representation in the proposed iDF-SLAM. The architecture of the MLP is shown in Fig. \ref{fig:mapper}.
We choose the MLP with 8 hidden layers of feature size 256 and 2 output heads. 
Unlike the MLP used in the iMAP, we take the idea from the recent neural RGB-D reconstruction work \cite{neuralrgbd} and change one of the output heads from predicting the volume density to the T-SDF $s$. 
The input to the MLP is the 3-D position $\bm{p}_i$ of a given point. We also include the view directions $\bm{d}$ in the input to get better per-frame rendering outputs. The periodical positional embedding proposed in the NeRF \cite{nerf} is applied to both the 3-D position and the view direction to project the inputs into a higher space.

In the neural implicit mapper, both the pose of the current keyframe and the MLP are optimised.
We choose the alternative optimisation approach mainly because it converges faster when most of the visible map is well constructed.

\subsection{MLP Optimisation}
To optimise the MLP, we first sample a batch of $B$ pixels from a given keyframe. With a ray being shot from the camera centre through every sampled pixel $\bm{p}$, we sample another $S_{\bm{p}}$ points per ray, forming the input to the MLP. Furthermore, to make the network focus on updating the unseen parts of the map or the parts where it is forgetting, we adopt the active sampling from iMAP \cite{imap}, which divided the keyframe into $8 \times 8$ grids and sampled more points from regions with higher losses.

When training the MLP, we create supervision by comparing the rendered RGB-D frame with the original input, just like other NeRF-related networks \cite{nerf,neuralrgbd}.
Specifically, a photometric loss $\mathcal{L}_p$ can be calculated as the $L-2$ norm of the differences between the predicted RGB values $I_{\bm{p}}$ and the input values $\hat{I}_{\bm{p}}$.
\begin{equation}
    \mathcal{L}_p = \frac{1}{|B|}\sum_{\bm{p} \in B} || I_{\bm{p}} - \hat{I}_{\bm{p}} ||_2
\label{eqn:lossp}
\end{equation}

Regarding the geometric loss, we use the same loss as \cite{neuralrgbd}, which is a weighted sum of a free space loss and a truncation loss.
\begin{equation}
\begin{split}
    \mathcal{L}_g &= w_{fs}\mathcal{L}_{fs} + w_{tr}\mathcal{L}_{tr} \\
    \mathcal{L}_{fs} &= \frac{1}{|B|}\sum_{\bm{p} \in B} \frac{1}{|S^{fs}_{\bm{p}}|}\sum_{\bm{s} \in S^{fs}_{\bm{p}}} || T_{\bm{s}} - tr ||_2 \\
    \mathcal{L}_{tr} &= \frac{1}{|B|}\sum_{\bm{p} \in B} \frac{1}{|S^{tr}_{\bm{p}}|}\sum_{\bm{s} \in S^{tr}_{\bm{p}}} || T_{\bm{s}} - \hat{T}_{\bm{s}} ||_2 \\
\end{split}
\end{equation}
where $tr$ is the truncation distance, $T_{\bm{s}}$ is the predicted T-SDF value at a point $\bm{s}$ and $\hat{T}_{\bm{s}}$ is the ground truth T-SDF value computed from the input depth measurement $\hat{D}_{\bm{p}}$. $S^{fs}_{\bm{p}}$ is the subset of the sampled points along the ray with points that fall between the camera centre and the near truncation region. Similarly, $S^{tr}_{\bm{p}}$ is the subset whose points fall into the truncation regions of the surface. With the free-space and the truncation regions explicitly formulated in the loss function, we ensure the network learning where the space is unoccupied.

Assuming that the parameters of the MLP are $\theta_{mlp}$, the final loss function to be optimised becomes:
\begin{equation}
    \underset{\theta_{mlp}}{\text{argmin}} \; \mathcal{L}_g + w_p \mathcal{L}_p 
\end{equation}

\subsection{Pose Optimisation}
The pose of the new keyframe is optimised in a similar manner as iNeRF \cite{inerf}.
We use exactly the same loss function as the MLP optimisation, except that the MLP parameters $\theta_{mlp}$ are fixed and only the pose parameter $\delta\bm{T}_c$ is updated. The resulting objective function for optimisation is:
\begin{equation}
    \underset{\delta\bm{T}_c}{\text{argmin}} \; \mathcal{L}_g + w_p \mathcal{L}_p 
\end{equation}
Additionally, when performing pose optimisation, we want the network to focus on the points corresponding to well-constructed parts of the map. Hence, we invert the active sampling mentioned above in pose optimisation. The keyframe is also divided into $8 \times 8$ grids, but instead of sampling more points from regions with higher losses, we now sample more points from regions with lower losses.

\subsection{Covisible Graph and Keyframe Culling}
Since the MLP is trained with input fed in sequentially, \textit{i.e.} in the continual learning setup, catastrophic forgetting happens. To prevent this, we follow iMAP \cite{imap} and adopt the replay-based approach, which maintains a buffer of keyframes and replays them throughout the training process. However, unlike iMAP which uses information gain based on predicted depth as the criteria to determine which keyframe to replay,
we propose to use the covisibility score between each pair of keyframes as we found that information gain in iMAP is computationally expensive and not very reliable with the T-SDF map, especially when the MLP has a good hallucination about the unseen parts of the map. In addition, since we create new keyframes whenever the neural tracker is lost or there is enough movement in the camera motion, we end up with more keyframes than iMAP, making keyframe culling necessary. 

Specifically, given a new keyframe $KF_n$ with optimised pose $\bm{T}_{n}$ and a set of previously stored keyframes $\mathbb{F} = \{KF_i\}$ and associated poses $\mathbb{T} = \{\bm{T}_{i}\}$ , we first compute the relative transformation from the new keyframe to every other keyframes $\bm{T}^i_{n}$. Then using the intrinsic matrix $\bm{K}$ and depth measurements, we can project every pixel in the current keyframe to a previously stored keyframe $KF_i$ with:
\begin{equation}
    \bm{u}_i = \bm{K} \bm{T}^i_{n} \bm{K}^{-1} \bm{u}_n d_n / d_i
\end{equation}
By checking whether the projected pixel $\bm{u}_i$ falls within the valid boundary of the keyframe, we can compute the covisibility score by dividing the number of valid projected pixels by the total number of pixels in the keyframe.

If the covisibility score of the keyframe $KF_i$ is greater than a predefined threshold $\sigma_{cull}$, we remove the new keyframe, as it is largely overlapped with the previously stored keyframe. Otherwise, the covisibility score is checked against another lower threshold $\sigma_{covis}$. If the score is greater than this new threshold, the new keyframe $KF_n$ is connected to the keyframe $KF_i$ in the covisible graph. 

When choosing the keyframes to replay, we randomly select a maximum of $N_{rep}$ keyframes that are not connected in the covisible graph. We also enforce that the new keyframe is always selected in every optimisation iterations.

\section{Neural Tracker}

The modified unsupervised R\&R (URR) model \cite{urr} is adopted as the neural tracker. Though any point cloud registration networks could be used in our iDF-SLAM, we favour the URR model mainly because it is trained and can be finetuned without ground truth supervision,
thus avoiding introducing ground-truth poses throughout our pipeline.

Given an input RGB-D frame and a reference RGB-D keyframe, the model first extracts feature maps from the RGB channels of both frames with a fully convolutional feature extractor. 
Then, using the depth channel, we can obtain a pair of point clouds by back-projection. 
Since neither striding nor pooling is used in the extractor, the output feature maps have the same height and width as the input images. Therefore, each point in the point cloud is associated with a unique feature vector. By conducting a \textit{k}-nearest neighbours (kNN) search in the feature space, a set of correspondences can be found between the two point clouds. 
Finally, the relative transformation from the current frame to the reference keyframe is computed using a weighted Procrustes algorithm \cite{choy2020}.

\begin{figure}[t]
    \centering
    \includegraphics[width=0.5\textwidth]{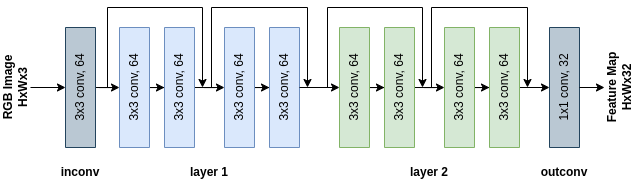}
    \caption{The architecture of the fully convolutional feature extractor used in the neural tracker.}
    \label{fig:tracker}
    \vspace{-3ex}
\end{figure}

The architecture of the feature extractor is shown in Fig. \ref{fig:tracker} where layer 1 and layer 2 share the same architecture and are the first convolutional block of ResNet18 \cite{resnet}.
We follow \cite{urr} to pretrain the extractor on the ScanNet \cite{scannet} dataset. For detailed pretraining procedure, we refer the readers to their paper.
To make the pretrained model more robust in the odometry task, we make the following modifications.

\subsection{Correspondences Selection}

\begin{figure}[]
\centerline{
\begin{tabular}{c}
     \includegraphics[width=0.45\textwidth]{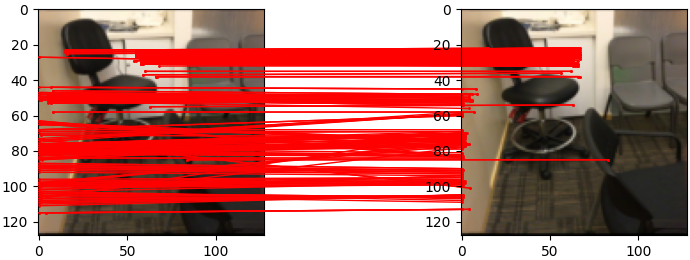}\\ 
     \includegraphics[width=0.45\textwidth]{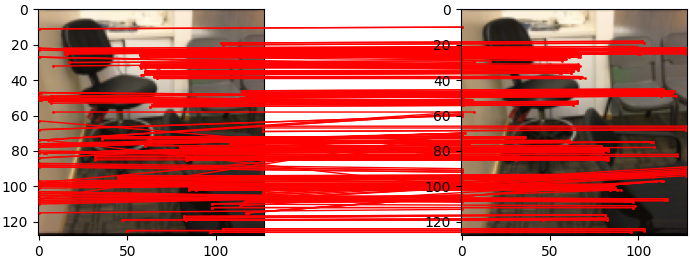}\\
\end{tabular}
}
\caption{Examples of matched feature points for pose estimation. (Top) Correspondences found with the default URR model, of which the feature points are clustered together in only certain regions of the input images; (Bottom) Correspondences found with our modified URR model, and the resulting feature points are spread out in the entire input images.}
\label{fig:features_points}
\vspace{-3ex}
\end{figure}

In the default URR model, the top-$K$ feature points are selected with the kNN search in the feature space. However, we notice that the correspondences found with the default URR model tend to be close to each other, resulting in the selected feature points for later pose estimation only coming from a small region of the input image; examples are given in Fig. \ref{fig:features_points} (Top). Due to the noise in the depth measurement and the inevitable outliers in the correspondences, such a set of feature points will likely make the estimated pose drift in a wrong orientation but stay undiscovered.

To solve this problem, we take the idea from ORB-SLAM \cite{orbslam} to enforce that feature points are selected from different regions of the input images, thus ensuring a homogeneous distribution. Specifically, we divide the input images into $4\times 4$ grids and set a maximum number of correspondences that can be found per grid. By doing so, the resulting set of correspondences is spread out in the input image, leading to more robust pose estimation in the odometry task.
Examples are given in Fig. \ref{fig:features_points} (Bottom). In practise, this maximum number is chosen as $K/16$.

\subsection{Scene Specific Features}
To enable the feature extractor to find the best features to estimate the camera pose in a specific scene, we propose to finetune the extractor using all previously stored keyframes. 
Once there are at least $N_{kf}$ keyframes stored in the system, we fix the weights of the mapper network $\theta_{mlp}$ and finetune the parameters of the feature extractor $\theta_{conv}$. Due to the limited number of keyframes can be generated from a finite-length sequence, we only update the weights in the outconv layer and fix the rest.

In each iteration, we first randomly select a pair of consecutive keyframes and pass the pair through our proposed iDF-SLAM to get the rendered RGB image and depth map. 
By comparing the rendered RGB image and the depth map with the original input,
a photometric loss and a geometric loss can be computed.
With the additional correspondence loss, we have the final loss of this finetuning process:
\begin{equation}
    \underset{\theta_{conv}}{\text{argmin}} \; w_p\mathcal{L}_p + w_d\mathcal{L}_d + w_r\mathcal{L}_{r}
\end{equation}
Specifically, the photometric loss is in the same formulation as in Eqn. \ref{eqn:lossp}. However, for the geometric loss, we directly compare the predicted depth values $D$ with the raw depth input $\hat{D}$.
\begin{equation}
    \mathcal{L}_d = \frac{1}{|B|}\sum_{\bm{p} \in B} || D_{\bm{p}} - \hat{D}_{\bm{p}} ||_2
\end{equation}
Finally, the correspondence loss is defined as the weighted sum of the Euclidean distance between the correspondences:
\begin{equation}
    \mathcal{L}_r = \frac{1}{|C|} \sum_{(c, r, w) \in C} w ||\bm{p}_r - \bm{T}_c \bm{p}_c||_2
\end{equation}
where $C = \{c, r, w\}$ is the set of correspondences with $c$, $r$ indexing the selected points of the current frame and the reference keyframe and $w$ being the confidence of the pair.


\begin{table*}
  \caption{Reconstruction Results of 8 Scenes in the Replica Dataset. Compared with iMAP and NICE-SLAM, our approach yields better results in most scenes.}
  \label{tab:replica}
  \vspace{-2ex}
  \scriptsize%
	\centering%
  \begin{tabu}{%
	l%
	l%
	*{2}{c}%
	*{2}{c}%
	*{2}{c}%
	*{2}{c}%
	*{2}{c}%
	*{2}{c}%
	*{2}{c}%
	*{2}{c}%
	*{2}{c}%
	}
  \toprule
  Methods & Metric & Room-0 & Room-1 & Room-2 & Office-0 & Office-1 & Office-2 & Office-3 & Office-4 & Avg. \\
  \midrule
  \multirow{3}{*}{NICE-SLAM\cite{niceslam}} 
  & \textbf{Acc.}[cm]$\downarrow$ 
  & 3.53 & \textbf{3.60} & \textbf{3.03} & 5.56 & \textbf{3.35} & 4.71 & 3.84 & 3.35 & 3.87 \\
  & \textbf{Comp.}[cm]$\downarrow$ 
  & 3.40 &  \textbf{3.62} & \textbf{3.27} & 4.55 & 4.03 & 3.94 & 3.99 & 4.15 & 3.87 \\
  & \textbf{Comp. Ratio}[$<5$cm \%]$\uparrow$ 
  & 86.05 & 80.75 & \textbf{87.23} & 79.34 & 82.13 & 80.35 & 80.55 & 82.88 & 82.41 \\
  
  \midrule 
  \multirow{3}{*}{iMap\cite{imap}} 
  & \textbf{Acc.}[cm]$\downarrow$ 
  & 3.58 & 3.69 & 4.68 & 5.87 & 3.71 & 4.81 & 4.27 & 4.83 & 4.43 \\
  & \textbf{Comp.}[cm]$\downarrow$ 
  & 5.06 & 4.87 & 5.51 & 6.11 & 5.26 & 5.65 & 5.45 & 6.59 & 5.56 \\
  & \textbf{Comp. Ratio}[$<5$cm \%]$\uparrow$ 
  & 83.91 & \textbf{83.45} & 75.53 & 77.71 & 79.64 & 77.22 & 77.34 & 77.63 & 79.06 \\
  
  \midrule 
  \multirow{3}{*}{iMap*\cite{imap}} 
  & \textbf{Acc.}[cm]$\downarrow$ 
  & 4.07 & 3.86 & 5.17 & 5.40 & 4.04 & 5.23 & 4.30 & 4.98 & 4.63 \\
  & \textbf{Comp.}[cm]$\downarrow$ 
  & 4.73 & 4.32 & 5.53 & 4.95 & 5.27 & 5.40 & 4.94 & 5.08 & 5.03 \\
  & \textbf{Comp. Ratio}[$<5$cm \%]$\uparrow$ 
  & 79.12 & 76.21 & 69.19 & 77.47 & 76.70 & 70.53 & 73.51 & 71.81 & 74.32 \\
  
  
  \midrule
  \multirow{3}{*}{\textbf{iDF-SLAM} } & 
  \textbf{Acc.}[cm]$\downarrow$ 
  & \textbf{2.30} & 10.25 & 4.43 & \textbf{5.01} & 7.38 & \textbf{3.08} & \textbf{3.12} & \textbf{3.32} & 4.86 \\
  & \textbf{Comp.}[cm]$\downarrow$ 
  & \textbf{1.39} & 7.31 & 3.64 & \textbf{2.70} & \textbf{1.70} & \textbf{2.76} & \textbf{1.88} & \textbf{2.46} & \textbf{3.05} \\
  & \textbf{Comp. Ratio}[$<5$cm \%]$\uparrow$ 
  & \textbf{95.68} & 61.60 & 78.37 & \textbf{88.54} & \textbf{94.57} & \textbf{88.51} & \textbf{92.05} & \textbf{91.36} & \textbf{86.34} \\
  \midrule
  \end{tabu}%
  \vspace{-3ex}
\end{table*}

\section{Experiments}

\begin{figure*}[t]
\centering
\includegraphics[width=\textwidth]{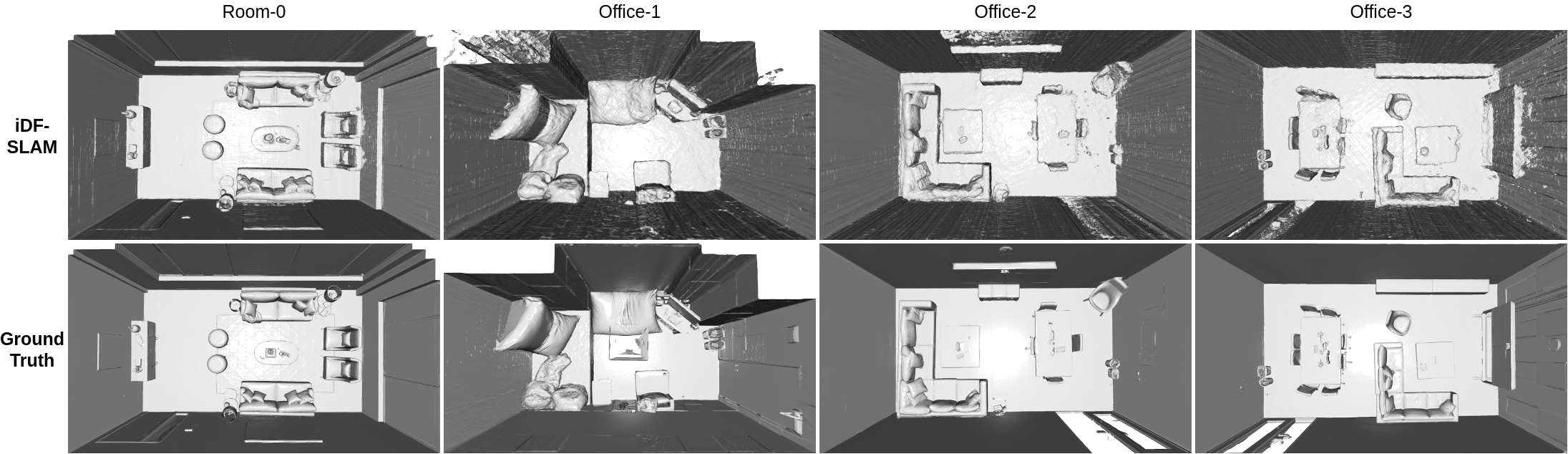}
\caption{
Examples of the Replica reconstructions. The reconstructions of the proposed iDF-SLAM are in the top row and ground truth reconstructions are in the bottom row.
}
\label{fig:recon-example}
\vspace*{-2ex}
\end{figure*}

\subsection{Implementation Details}

For all of our experiments, we sample $B=1024$ pixels per image with $S_{\bm{p}}=144$ 
points sampled along each ray. 
We set the truncation distance to $tr =10 \,cm$ and the maximum number of keyframes to be replayed to $N_{rep}=10$.
As for keyframe culling and covisible graph construction, we set $\sigma_{cull} = 0.5$ and $\sigma_{covis} = 0.3$. 
Finally, in neural tracker, we select $K=200$ pairs of correspondences for pose estimation and require at least $N_{kf}=10$ before starting finetuning.

As for optimisations, we use the Adam \cite{adam} optimiser for all three optimisation tasks though different learning rates are chosen. When performing MLP optimisation, the learning rate is set to $\eta_{mlp} = 0.005$ with no learning rate decay. For pose optimisation, the initial learning rate is also set to $\eta_{pose} = 0.005$, but a step scheduler is also implemented to decay the learning rate by 0.7 every 10 iterations. Finally, the learning rate for finetuning the feature extractor is fixed to $\eta_{conv} = 0.0001$.

\subsection{Datasets and Metrics}
We test the performance of the proposed iDF-SLAM in scenes from a virtual dataset, Replica \cite{replica}, and another real dataset, ScanNet \cite{scannet}.
Specifically, we evaluate the reconstruction quality only on the Replica dataset, while the tracking performance on both datasets. Following iMAP \cite{imap} and NICE-SLAM \cite{niceslam}, we use the same 2000-frame long RGB-D sequences for 8 scenes in the Replica dataset and the same 6 scenes from the ScanNet dataset.

In terms of evaluation metrics, we follow iMAP and NICE-SLAM and use three different metrics for scene reconstruction evaluation: \textit{accuracy}, \textit{completion} and \textit{completion ratio}. When evaluating the estimated camera trajectory, \textit{absolute trajectory error} (ATE) is adopted. We refer the readers to the iMAP paper \cite{imap} for more details on the evaluation metrics.

\subsection{Evaluations}
We compare the proposed iDF-SLAM with two recent NeRF-based SLAM systems, iMAP \cite{imap} and NICE-SLAM \cite{niceslam}. However, since the official implementation of iMAP has not yet been released, we use the implementation from the authors of NICE-SLAM for the results not reported in the iMAP paper. The results from the re-implemented iMAP is denoted as iMAP* in the tables.

\textbf{Runtime and Memory}: We perform the experiments on a single GeForce RTX 2080 TI GPU. Our system only consumes around 4GB VRAM. For a 2000-frame long sequence, our iDF-SLAM takes about 7 hours to finish, while in comparison, a pure optimisation-based system like iMAP takes about 12 hours to run when using the same number of optimisation loops.

\begin{table*}[t]
  \caption{Camera Tracking Results on the Replica Dataset.}
  \label{tab:replica_ate}
  \vspace{-2ex}
  \scriptsize%
	\centering%
  \begin{tabu}{%
	l%
	l%
	*{2}{c}%
	*{2}{c}%
	*{2}{c}%
	*{2}{c}%
	*{2}{c}%
	*{2}{c}%
	*{2}{c}%
	*{2}{c}%
	*{2}{c}%
	}
  \toprule
  Methods & Metric & Room-0 & Room-1 & Room-2 & Office-0 & Office-1 & Office-2 & Office-3 & Office-4 & Avg. \\
    

  \midrule
  \multirow{3}{*}{NICE-SLAM \cite{niceslam}} 
  & \textbf{RMSE}[m]$\downarrow$ 
  & \textbf{0.017} & 0.020 & \textbf{0.016} & \textbf{0.010} & \textbf{0.009} & 0.014 & 0.040 & 0.031 & \textbf{0.020} \\
  & \textbf{mean}[m]$\downarrow$ 
  & \textbf{0.015} & 0.018 & \textbf{0.012} & \textbf{0.009} & 0.008 & 0.012 & 0.021 & 0.021 & \textbf{0.014} \\
  & \textbf{median}[m]$\downarrow$ 
  & \textbf{0.014} & 0.017 & \textbf{0.010} & \textbf{0.008} & 0.007 & 0.011 & \textbf{0.013} & 0.015 & \textbf{0.012} \\

  \midrule 
  \multirow{3}{*}{iMAP* \cite{imap}} & \textbf{RMSE}[m]$\downarrow$ 
  & 0.701 & 0.045 &0.022 & 0.023 & 0.017 & 0.049 & 0.584 & 0.026 & 0.183 \\
  & \textbf{mean}[m]$\downarrow$ 
  & 0.589 & 0.040 & 0.020 & 0.017 & 0.016 & 0.032  & 0.549  & 0.022  & 0.160 \\
  & \textbf{median}[m]$\downarrow$ 
  & 0.448 & 0.034 & 0.017 & 0.014 & 0.014 & 0.024  & 0.476  & 0.019  & 0.130 \\

  \midrule
  \multirow{3}{*}{\textbf{iDF-SLAM}} 
  & \textbf{RMSE}[m]$\downarrow$ 
  & 0.018 & 0.059 & 0.026 & 0.016 & 0.021 & 0.018 & \textbf{0.019} & \textbf{0.021}  & 0.025 \\
  & \textbf{mean}[m]$\downarrow$
  & \textbf{0.015} & 0.049 & 0.025 & 0.013 & 0.016 & 0.015 & \textbf{0.015} & \textbf{0.017}  & 0.020 \\
  & \textbf{median}[m]$\downarrow$
  & \textbf{0.014} & 0.042 & 0.024 & 0.010 & 0.014 & 0.012 & \textbf{0.013} & \textbf{0.014}  & 0.018 \\

  \midrule
  \end{tabu}%
  \vspace{-3ex}
\end{table*}

\begin{table}[t]
    \centering
    \caption{Camera Tracking Results on ScanNet. ATE RMSE [cm] is used as the evaluation metric.}
    \label{tab:scannet_ate}
    \vspace{-2ex}
    \begin{tabu}{%
	l%
	*{2}{c}%
	*{2}{c}%
	*{2}{c}%
	*{2}{c}%
	*{2}{c}%
	*{2}{c}%
	*{2}{c}%
	}
    \toprule
    Scene ID  & 0000  & 0059 & 0106 & 0169 & 0181 & 0207 
    \\
    \midrule
    NICE-SLAM \cite{niceslam} & \textbf{11.3}  & 12.0 & 9.0  & \textbf{12.0} & \textbf{11.2} & \textbf{12.8} 
    \\
    iMAP*  \cite{imap} & 197.1 & 18.9 & 19.0 & 96.4 & 37.3 & 28.7 
    \\
    \textbf{iDF-SLAM}       &  57.7 & \textbf{8.2} & \textbf{5.8} 
        & 39.9 &  29.1 
        &  15.4 
        \\
    \midrule
    \end{tabu}%
    \vspace{-3ex}
\end{table}


\textbf{Reconstruction}: In Table \ref{tab:replica}, we present the quantitative reconstruction results on the Replica dataset. Our proposed iDF-SLAM outperforms both NICE-SLAM and iMAP in most of the scenes to a large extent. Especially for the completion and the completion ratio where a consistent improvement of 2 cm and 10\% can be observed. Although the reconstruction quality drops significantly in Room-1 and Room-2, causing the average reconstruction accuracy slightly higher than iMAP and NICE-SLAM, the average completion and completion ration are still better than iMAP and NICE-SLAM. For qualitative results, we also provide examples of the reconstructed Replica scenes in Fig. \ref{fig:recon-example} along with the ground truth reconstructions.

\textbf{Tracking}: In Tables \ref{tab:replica_ate} and \ref{tab:scannet_ate}, we present the results of the camera tracking. When testing on the Replica dataset, the proposed iDF-SLAM outperforms iMAP and NICE-SLAM in Room-0, Office-3 and Office-4 and demonstrates competitive performance in Office-0 and Office-2. Moving on the real scenes, \textit{i.e.} the ScanNet dataset, the proposed iDF-SLAM exhibits better tracking accuracy than NICE-SLAM and iMAP in scene0059 and scene 0106. Though the tracking performance is not as good as NICE-SLAM in the rest of the scenes, it is still much better than iMAP.


\subsection{Discussions}

It is worth noting that our iDF-SLAM has better reconstruction completion than NICE-SLAM and iMAP on Replica Office-1, while worse reconstruction accuracy and tracking performance. We believe that the reason for this is two-fold. First of all, because of the nature of the T-SDF map representation, the inobservable regions behind the wall contain random noises which lead to noisy surfaces in the reconstruction. In addition, Office-1 has irregular walls, those noises are difficult to remove during evaluation, hence a worse reconstruction accuracy.

As for the tracking performance, we believe that it is because in the iDF-SLAM, not all frames are optimised with respect to the map to obtain better pose estimation. We only perform pose optimisation and update map on keyframes. Besides, the neural tracker we used in the front-end is a feature-based tracker, which is more sensitive to drastic motions and blurs of the input images. Therefore, although the keyframes in Office-1 can be optimised to a good estimates, hence good reconstruction result, the non-keyframes causes the overall camera tracking results to be worse than NICE-SLAM and iMAP.

The same effects can also be observed from the results of Replica Room-1 and Room-2. Especially in Replica Room-1, the radical rotation motion in the camera movement makes it extremely difficult for the neural tracker to work properly. And because the map does not update per frame, the lost camera cannot be optimised to a good pose estimate, leading to a significant drop in both the reconstruction and camera tracking performance.

In addition to the drastic camera motions, another limitation of the proposed iDF-SLAM is the notorious long-term drift, which is even exaggerated by the feature-based front-end. This is proved by the notable drop in tracking performance in the long sequences of ScanNet data. Furthermore, due to the holes in the real depth measurements and the relatively lower map update frequency, we are unable to achieve a tracking performance as good as NICE-SLAM on the ScanNet dataset.

\section{Conclusions}

In conclusion, we present iDF-SLAM, which has a feature-based front-end for camera tracking using deep features extracted by a neural network and a NeRF-style network to implicitly represent the map. The system can be trained in an online, self-supervised and end-to-end fashion. Compared to the other two recent NeRF-based SLAM systems, the proposed iDF-SLAM achieves SOTA reconstruction results and exhibits competitive tracking accuracy. In future work, we intend to implement loop closure to improve the system's performance in larger environments.









\end{document}